\documentclass{article}

\usepackage{arxiv}
\usepackage[utf8]{inputenc} % allow utf-8 input
\usepackage[T1]{fontenc}    % use 8-bit T1 fonts
\usepackage{hyperref}       % hyperlinks
\usepackage{url}            % simple URL typesetting
\usepackage{booktabs}       % professional-quality tables
\usepackage{amsfonts}       % blackboard math symbols
\usepackage{nicefrac}       % compact symbols for 1/2, etc.
\usepackage{microtype}      % microtypography
\usepackage{lipsum}
\usepackage{graphicx}
\usepackage{natbib}
\usepackage{amsmath}
\usepackage{color,soul}
\usepackage{underscore}
\graphicspath{ {./images/}}
\usepackage{float}
\usepackage[svgnames]{xcolor}  % or x11names
\usepackage{soul}
\usepackage[nonumberlist,acronym]{glossaries}
\makeglossaries

\usepackage{pmboxdraw}
\DeclareUnicodeCharacter{2581}{\pmboxdrawuni{2581}}

\title{\textsc{Visualizing LLM Latent Space Geometry Through Dimensionality Reduction}}

\author{
  Alex Ning\thanks{Equal contribution} \\
  Department of Computer Science\\ 
  University of Virginia\\
  \texttt{rnx2bc@virginia.edu} \\
   \And
  Vainateya Rangaraju\footnotemark[1] \\
  Department of Computer Science\\
  University of Virginia\\
  \texttt{prr3gw@virginia.edu} \\
  \And
 Yen-Ling Kuo\\
  Department of Computer Science\\
  University of Virginia\\
  \texttt{ylkuo@virginia.edu} \\
}

\begin{document}
\maketitle
\begin{abstract}
Large language models (LLMs) achieve state-of-the-art results across many natural language tasks, but their internal mechanisms remain difficult to interpret. In this work, we extract, process, and visualize latent state geometries in Transformer-based language models through dimensionality reduction. We capture layerwise activations at multiple points within Transformer blocks and enable systematic analysis through Principal Component Analysis (PCA) and Uniform Manifold Approximation and Projection (UMAP). We demonstrate experiments on GPT-2 and LLaMa models, where we uncover interesting geometric patterns in latent space. Notably, we identify a clear separation between attention and MLP component outputs across intermediate layers, a pattern not documented in prior work to our knowledge. We also characterize the high norm of latent states at the initial sequence position and visualize the layerwise evolution of latent states. Additionally, we demonstrate the high-dimensional helical structure of GPT-2's positional embeddings and the sequence-wise geometric patterns in LLaMa. We make our code available at \url{https://github.com/Vainateya/Feature_Geometry_Visualization}. A better formatted blog-post with identical content is available at \url{https://iclr-blogposts.github.io/2026/blog/2026/vis-llm-latent-geometry/}.
\end{abstract}

% keywords can be removed

\section{Introduction}
\label{sec:introduction}
Despite enormous advances made in the field of machine learning (ML) research, understanding a model's internal decision-making processes remains a difficult challenge. Model interpretability is central to areas such as alignment and explainable AI, and the field is burgeoning with an influx of work. However, many foundational questions are still unresolved. Against this backdrop, mechanistic interpretability has emerged as a field that aims to achieve a granular, causal understanding of neural networks, particularly large language models (LLMs), by reverse engineering their internal components. One promising avenue for understanding LLMs is analyzing their representations. Feature geometry, the structure and organization of representations within high-dimensional latent space, offers a way to study how abstract features are encoded and transformed across model layers. By examining geometric relationships between latent states, such as directions, clusters, and manifolds, researchers can gain insight into how LLMs generalize, reason, and abstract.

In this work, we analyze the feature geometry of LLMs by capturing latent states across multiple components and projecting them into interpretable low-dimensional spaces using PCA and UMAP. This approach allows visualizations of how representations evolve through the Transformer model. Throughout this work, we aim to provide a strong background and exposition for better accessibility. To ensure clarity, a glossary of key technical terms used throughout this paper is provided in section~\ref{sec:glossary}.

\section{Background}

\subsection{Transformers}
\label{subsec:Transformers}

\begin{figure}[H]
    \centering
    \setlength{\fboxsep}{0pt}
    \includegraphics[width=\textwidth]{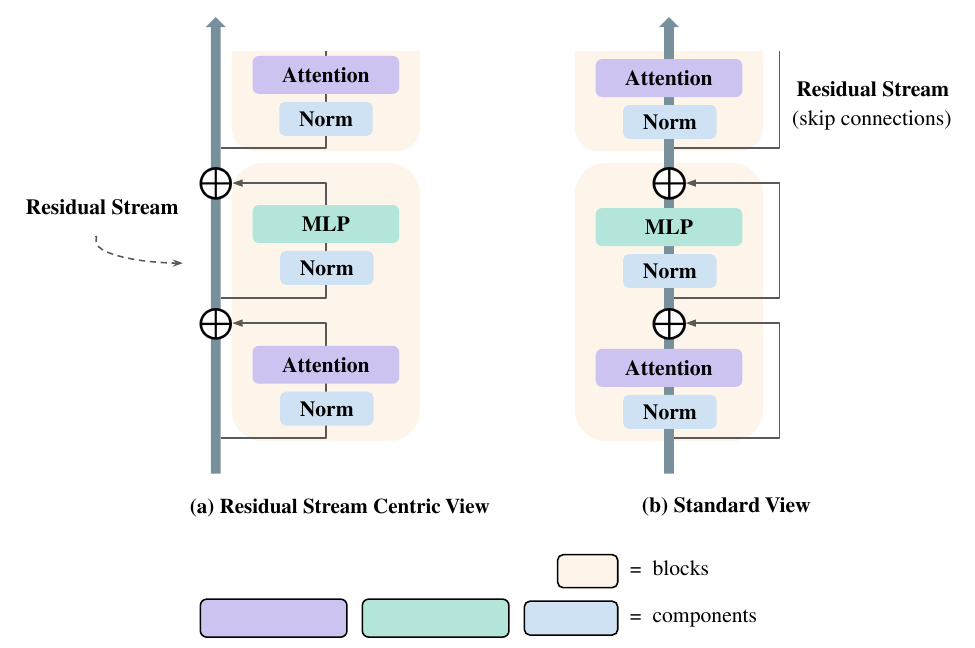}
   \caption{The two equivalent perspectives on the Transformer architecture}
    \label{fig:residual_views}
\end{figure}

Originally introduced by Vaswani et al. in 2017 \citep{vaswani2023attentionneed}, Transformers have achieved great recognition for their state-of-the-art performance across most language modeling tasks. Despite this success, their internal mechanisms remain difficult to understand, motivating extensive work in mechanistic interpretability (see Section \ref{sec:introduction}). Since our experiments focus on analyzing decoder-only Transformers (e.g., GPT-2~\citep{radford2019language} and LLaMa~\citep{touvron2023llamaopenefficientfoundation}), we begin by providing a brief overview of the Transformer architecture. To ensure clarity amongst various terminologies in recent literature, we also establish a consistent set of terms that will be used throughout this paper. That is, we formalize a decoder-only Transformer as a sequence of blocks, each consisting of four primary components/layers: the normalization layer (e.g., layer norm or RMSNorm) preceding the attention component, the multi-head self-attention, the normalization layer preceding the MLP (multilayer perceptron), and the MLP. Although the original Transformer architecture utilized a post-norm design, where the normalization layer came after the attention or MLP component, most modern architectures (e.g. GPT-2 and LLaMa) adopt a pre-norm design. Thus, we depict the pre-norm architecture in figures \ref{fig:residual_views} and \ref{fig:latent_capture}. We use the term ``final norm'' to refer to the normalization applied after the last Transformer block. This is immediately followed by the unembedding layer, which maps the final latent representation into vocabulary logits for token prediction. Finally, we use 0-based indexing when referring to blocks, layers, and sequence positions throughout the paper. For example, ``block 0'' denotes the \textit{first} block in the Transformer, and ``sequence position 0'' denotes the \textit{first} token position in a sequence. In order to avoid ambiguity between ``first'' and index ``1'', we will refer to the ``first'' ($0$-th) item as the ``initial'' item. While Transformers are often conceptualized as a sequential operation through each of their components with residual connections between layers, we adopt the mathematically equivalent perspective that these skip connections collectively form the central communication channel through which all components interact. We refer to this pathway as the residual stream, as formalized by Elhage et al. in 2021~\citep{elhage2021mathematical}.

Figure \ref{fig:residual_views} illustrates these differing perspectives. In this residual stream-centric view, all components of a Transformer (the token embedding, attention heads, MLP layers, and unembedding layer) communicate with each other by reading and writing to different subspaces of the residual stream. Rather than thinking of information flowing sequentially through layers, we conceptualize each layer as reading its input from the residual stream (by performing a linear projection), and then writing its result to the residual stream by adding a linear projection back in \citep{elhage2021mathematical}. This additive structure preserves a path for an identity pathway to flow through the model and into each component, unobstructed by any operation other than the direct sum of component updates back into the residual stream. We highlight the importance of the residual stream perspective for two reasons. Firstly, it allows us to treat all components of the Transformer as operating within a shared representation space, intuitively illustrating the collaborative nature of component interactions. This shared space enables different components to develop a collective understanding of features and their corresponding directions, making it possible to interpret representational changes as coordinated rather than isolated transformations. Secondly, this framing provides a conceptual basis for analyzing how the geometry of these shared representations evolves throughout the residual stream, thereby offering insights into how components jointly shape the model's feature space.

\subsection{Linear Representation Hypothesis}
\label{subsec:lrh}
LLMs learn internal features, which are abstract properties or characteristics of data that guide their predictions and generalization. In language modeling, such features often correspond to linguistic attributes such as gender, tense, or sentiment. Mikolov et al. in 2013~\citep{mikolov-etal-2013-linguistic} showed that word embeddings capture these features through simple vector arithmetic. For example, $\langle \text{King} \rangle - \langle \text{Man} \rangle + \langle \text{Woman} \rangle \approx \langle \text{Queen} \rangle$, where angle brackets denote embedding vectors. This provided the first evidence that semantic and syntactic relations can be represented as linear patterns in latent space. This empirical finding has since been formalized in the context of large language models as the Linear Representation Hypothesis (LRH)~\citep{park2024linearrepresentationhypothesisgeometry}, which posits that high-level concepts are represented as directions (i.e., one-dimensional subspaces) in representation space. This implies an intuition where learned representations are approximately linear combinations of feature directions.

LRH is significant for mechanistic interpretability in two respects. First, LRH implies a property of decomposability, which conveniently states that complex representations can be analyzed in terms of independent, concept-aligned features. This property provides a tractable foundation for understanding LLM latent space dynamics. Second, under the assumption that features are represented linearly, dimensionality reduction methods can more reliably reveal the organization and combination of features within the latent space. In this framing, each feature $i$ may be represented by a unit vector $e_i$, with an input's representation expressed as $\sum x_i e_i$, where $x_i$ denotes the feature's strength or presence \citep{elhage2022superposition}. With that said, Linear Representation Hypothesis remains a \textit{hypothesis}, and there are limitations to the interpretations it can provide. Transformer models implement many behaviors which are extremely complex, not yet understood, and highly non-linear. Engels et al. in 2025 \cite{engels2025languagemodelfeaturesonedimensionally} explored features which are \textit{not} one-dimensionally linear. Additionally, the Superposition Hypothesis~\citep{elhage2022superposition} posits that in order to represent a large number of independent features in a space with limited dimensionality, neural networks permit a certain amount of interference between features by embedding them into \textit{almost}-orthogonal directions instead of fully orthogonal directions. Although still linear, this degree of interference potentially clouds the benefits of interpretability and decomposability offered by LRH.

\subsection{Dimensionality Reduction}
\label{subsec:dimreduct}
Modern neural networks generate latent representations that live in extremely high-dimensional spaces. While these representations contain rich structural information, they are difficult to interpret directly. Dimensionality reduction techniques serve as a bridge between high-dimensional representations and human visualization. The goal is to compress embeddings into lower dimensions while preserving the salient structure, e.g., variance, clusters, or manifold geometry, so that latent state behavior can be more easily examined. In this work, we study both Principal Component Analysis (PCA) and Uniform Manifold Approximation and Projection (UMAP) as the primary methods for dimensionality reduction and visualization.

PCA is a widely used method for linear dimensionality reduction. PCA operates by finding a new orthogonal basis that captures directions of maximal variance in the data \citep{shlens2014tutorialprincipalcomponentanalysis}. By projecting high-dimensional vectors onto these principal axes, PCA allows one to rank-order components by their explanatory power. The method is computationally efficient, analytically grounded in linear algebra through eigenvalue decomposition or singular value decomposition, and does not require hyperparameter tuning. However, PCA is limited to capturing linear relationships and assumes that variance corresponds to meaningful structure, an assumption that may fail in non-Gaussian or highly nonlinear datasets \citep{shlens2014tutorialprincipalcomponentanalysis}. Despite these caveats, PCA remains a powerful baseline, is widely used in our experiments, and provides a stable reference frame for comparing geometric patterns.

On the other hand, Uniform Manifold Approximation and Projection (UMAP), proposed by McInnes et al. in 2018~\citep{mcinnes2020umapuniformmanifoldapproximation}, provides a nonlinear alternative designed to preserve both local neighborhoods and some aspects of global manifold structure. Rather than identifying variance-maximizing axes, UMAP builds a graph-based representation of data as a topological manifold and then optimizes a low-dimensional embedding that preserves this structure. In practice, UMAP is effective at revealing clustering patterns in latent states, producing interpretable visualizations. We chose UMAP over other nonlinear alternatives such as t-SNE \citep{JMLR:v9:vandermaaten08a} because it better maintains global relationships while still revealing local clustering patterns. Its flexibility comes at the cost of tunable hyperparameters (e.g., number of neighbors, minimum distance) and a lack of linear structure, but it has become a widely adopted tool in machine learning for exploring latent structure.

\subsection{Positional Embeddings/Encodings}
\label{subsec:positionalencodings}
Transformers are inherently permutation-invariant, meaning that without explicit positional information, they cannot distinguish between tokens based solely on order. To address this, models incorporate positional information into token representations so that attention mechanisms can account for sequence structure. This information is typically introduced in one of two ways: through learned positional embeddings or functional positional encodings, each with distinct implications for how visualizations depict positional geometry in a model’s latent space.

In many early Transformer architectures, such as GPT-2 (used in our experiments), positional information is injected via learned positional embeddings. Each token position in the input sequence is associated with a learned vector that is added to its corresponding token embedding before entering the first Transformer block. Because these embeddings are learned during training, they often exhibit consistent geometric patterns that reflect the model's internal representation of positional order. Within the context of our work, learned positional embeddings provide a straightforward baseline for the geometry of position-dependent features.

On the other hand, the LLaMa model we use employs a form of functional positional encoding called Rotary Positional Encodings (RoPE). Proposed by Su et al. in 2021~\citep{su2023roformerenhancedtransformerrotary}, RoPE applies a position-dependent rotation to query and key vectors within each attention head rather than adding fixed position vectors to the initial embeddings. This rotation encodes relative position information directly into the attention mechanism. Conceptually, RoPE embeds position in the geometry of relationships between tokens, rather than as an explicit position embedding vector.

This distinction serves as a key motivation behind \textit{Effects of Sequence Position} experiments in section~\ref{sec:effects_of_sequence_position}.

\section{Methodology}

\begin{figure}[H]
    \centering
    \setlength{\fboxsep}{0pt}
    \includegraphics[width=\textwidth]{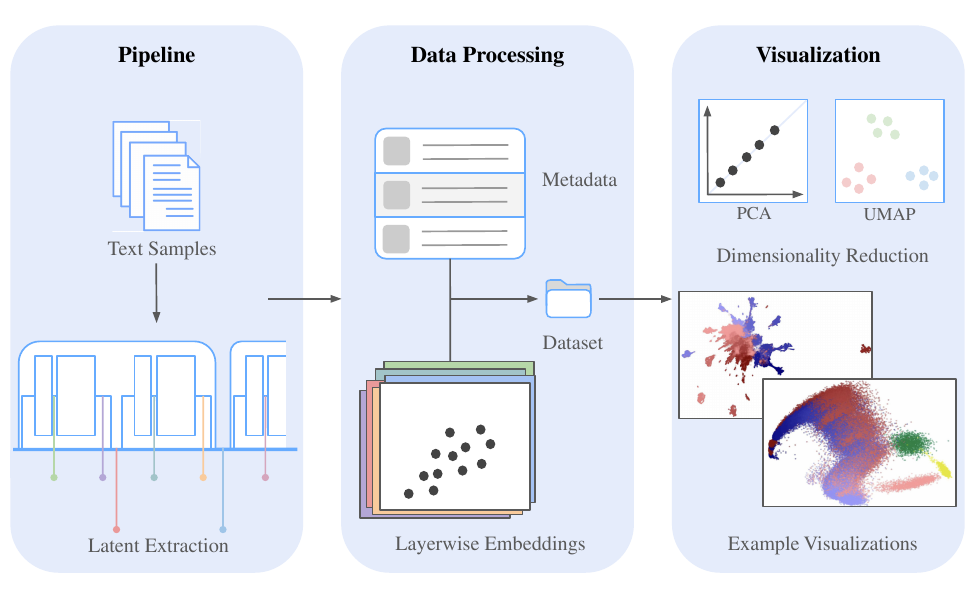}
   \caption{Overview of visualization pipeline: Text samples first pass through Transformer layers for latent extraction, these latent states are then organized into a structured dataset alongside metadata, then reduced via dimensionality reduction for interpretable visualizations.}
    \label{fig:system_diagram}
\end{figure}

The visualizations obtained from our dimensionality reduction experiments are the results of a 3-part process designed to capture, organize, and explore the internal representations of Transformer models. 
Figure \ref{fig:system_diagram} shows the overview of our methodology.
The first component is the pipeline, which handles the extraction of latent states by running models on text data and saving layerwise latent states alongside metadata and token information. Once this latent state dataset has been generated, the second component, data processing, provides tools for efficiently loading, filtering, and manipulating the stored representations. Finally, the third component focuses on dimensionality reduction and visualization, utilizing PCA and UMAP to project high-dimensional latent states into interpretable views. Together, these stages provide a unified workflow for moving from raw model activations to structured visualizations that highlight the geometric signatures present in the latent space.

\subsection{Pipeline} 
\label{subsec:pipeline}

\begin{figure}[H]
    \centering
    \setlength{\fboxsep}{0pt}
    \includegraphics[width=\textwidth]{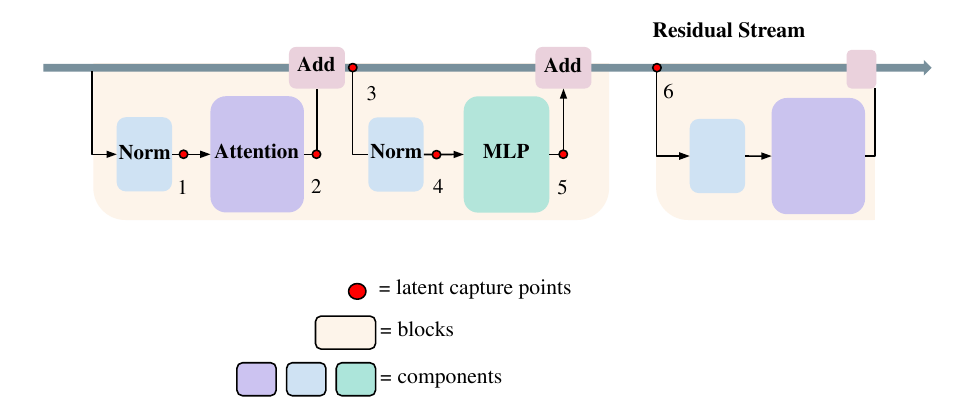}
   \caption{The six capture points within each Transformer block. Points 1 and 4 correspond to the outputs of the normalization layers (pre-attention and pre-MLP). Points 2 and 5 correspond to the outputs of the attention and MLP modules, respectively. Points 3 and 6 capture the residual stream after the attention and MLP additions.}
    \label{fig:latent_capture}
\end{figure}

The generation pipeline automates three stages: (1) generating inputs, (2) capturing latent states, and (3) saving labeled latent states for downstream processing and visualization. This design provides a reproducible and extensible foundation for our experiments (see section \ref{sec:results_and_discussion}). 

The pipeline supports two modes of operation. In the default ``text'' mode, text passages are sampled from a text dataset. For our experiments, we ran the model through the well-known Project Gutenberg (PG-19) dataset introduced by Rae et al. in 2019~\citep{rae2019compressivetransformerslongrangesequence}, consisting of a corpus of books written before 1919. Each passage is tokenized and truncated or padded to a fixed sequence length, ensuring uniformity across samples. In ``singular'' mode, by contrast, the pipeline samples individual tokens directly by iterating over the model's vocabulary. This allows for fine-grained analysis of token-level representations without contextual interference. Both modes are parameterized by the number of samples and the desired sequence length (set at $1$ for the singular mode).

Figure~\ref{fig:latent_capture} shows each of the capture points along the Transformer architecture. For each block, we record the outputs of each layer: both normalizations, the attention module, and the MLP. We denote the raw outputs of the attention and MLP components and their preceding norms as ``pre-add'' (points 1, 2, 4, and 5 in Figure~\ref{fig:latent_capture}), and the latent states after they have been added back into the residual stream as ``post-add'' (points 3 and 6 in Figure~\ref{fig:latent_capture}). In total, this yields six distinct captures per Transformer block.

Captured latent states of high dimensionality can optionally undergo post-processing to moderately reduce their dimensionality via PCA (e.g., from $4096$ to $512$ dimensions). This reduced dimensionality can substantially reduce memory usage and loading, processing, and visualization time with minimal impact on results.

\subsection{Dimensionality Reduction and Visualization}

Captured latent states are visualized in $2$D following dimensionality reduction. Not all captured latent states must be used for dimensionality reduction and visualization. Instead, a subset may be selected (e.g., only latent states from particular layers). For dimensionality reduction, both PCA and UMAP are used, with a GPU-accelerated implementation for UMAP (via cuML \citep{raschka2020machinelearningpythonmain}) utilized for visualizing large amounts of latent states. Before dimensionality reduction is performed, augmentations to the latent states can be applied. These include:

\begin{itemize}
    \item Converting latent states to unit length, which emphasizes directional structure over norm.
    \item Averaging the latent state vectors over any dimension(s) (sample dimension, sequence dimension, layer dimension). This may be done to remove the variance of a certain dimension to concentrate on the structure present in another.
\end{itemize}

PCA transformation is performed without mean centering to preserve the position of the origin.

We utilize a color-coding scheme to distinguish the layer and sequence dimensions. In our experiments, attention layers are blue while MLP layers are red. Darker shades indicate earlier layers while lighter shades indicate later ones. For the sequence dimension, we use a gradient to indicate earlier vs. later positions. 

Fitted dimensionality reduction models can be reused; PCA fitted on one set of latent states can be applied to a different set of latent states, and visualizations can preserve x- and y-axis limits between runs. This is especially useful for ablation studies, where subtle differences between experimental conditions must be compared in a consistent coordinate frame. Additionally, dimensionality reduction can be performed that reduces the number of dimensions to a number \textit{greater} than $2$. Then, any pair of the reduced dimensions can be visualized together in $2$D (e.g., dimensions $2$ and $5$ or $3$ and $7$). Figure \ref{fig:gpt2_pos_unit} shows the result of visualizing all possible pairs of dimensions after dimensionality reduction to $6$ dimensions (totaling ${6 \choose 2} = 15$ pairs), then stitching the resulting visualizations into a grid.

\section{Results and Discussion}
\label{sec:results_and_discussion}

\subsection{Experimental Setup}
\label{subsec:ExperimentalSetup}

We use GPT-2 Large ($774$M) and LLaMa-7B to generate the latent states. On all results using the PG-19 dataset, we use the following input shapes:

\begin{itemize}
    \item GPT-2: $128$ samples, each $1024$ tokens long
    \item LLaMa: $64$ samples, each $2048$ tokens long
\end{itemize}

We note that we use the maximum input length possible for both models. LLaMa is trained with a \texttt{<BOS>} (beginning of sentence) token prepended to all inputs. As a result, we prepend the \texttt{<BOS>} token to the beginning of all PG-19 inputs to LLaMa. After collecting the LLaMa latent states, we reduced the dimensionality of the latent states from $4096$ to $512$ using PCA. These dimensionality-reduced latent states were subsequently used in all visualizations. They were not used in any analyses of latent state norms; those used the original full dimensionality latent states for accurate results. All GPT-2 latent states were used in their full dimensionality.

Existing research demonstrates how different blocks in a Transformer have different roles. While earlier blocks focus on converting tokens into concepts as shown by Nanda et al. in 2023~\citep{nanda2023factfinding} and the final blocks heavily denoise and refine the output, intermediate (middle) blocks, which are more robust to swapping and deletion, gradually create and refine intermediate features in a shared representation space as demonstrated by Sun et al.~\citep{sun2025transformerlayerspainters} and Lad et al.~\citep{lad2025remarkablerobustnessllmsstages} in 2024. For some of our experiments, we are interested in investigating the behavior of the intermediate block latent states without interference from earlier or later blocks. As a result, we define here the specific ``intermediate'' blocks/layers which we will later refer to in our results and discussions for both our tested models. Because we could not find a clear methodology to define concrete boundaries for the "intermediate" blocks, the definitions we make are somewhat arbitrary. Nevertheless, it is important for us to have consistent definitions across our analyses. For GPT-2, we define these as the layers in blocks $2$-$8$ (from a range of $0$-$11$), while for LLaMa, they are the layers in blocks $6$-$27$ (from a range of $0$-$31$).

\subsection{Large Norm of 0-th Sequence Position Latent States}\label{sec:first_pos_large_norm}

\begin{figure}[H]
    \centering
    \setlength{\fboxsep}{0pt}
    \includegraphics[width=\textwidth]{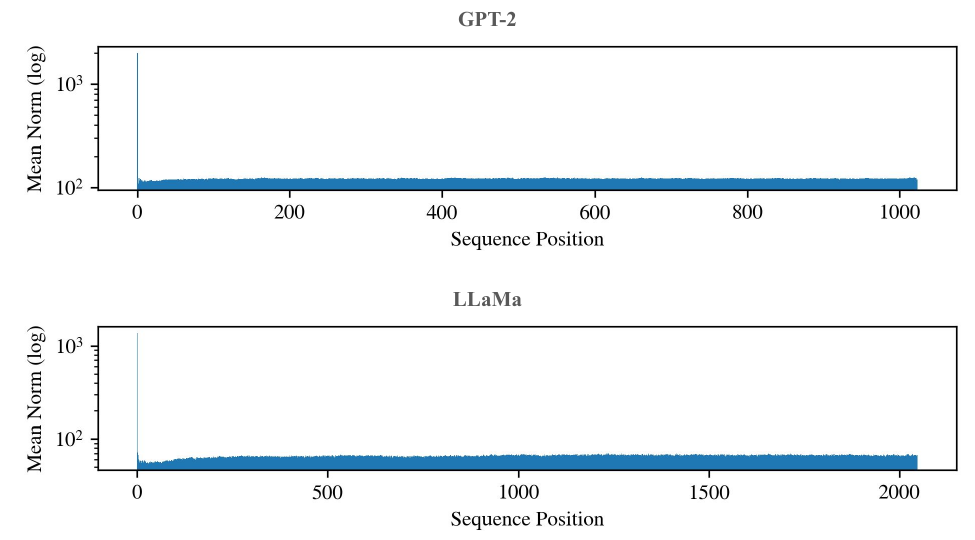}
    \caption{Norm of latent states from intermediate layers of both GPT-2 and LLaMa along sequence positions. Norms were averaged over both samples and layers. The dataset used is PG-19.}
    \label{fig:large_norm}
\end{figure}

\begin{figure}[H]
    \centering
    \setlength{\fboxsep}{0pt}
    \includegraphics[width=\textwidth]{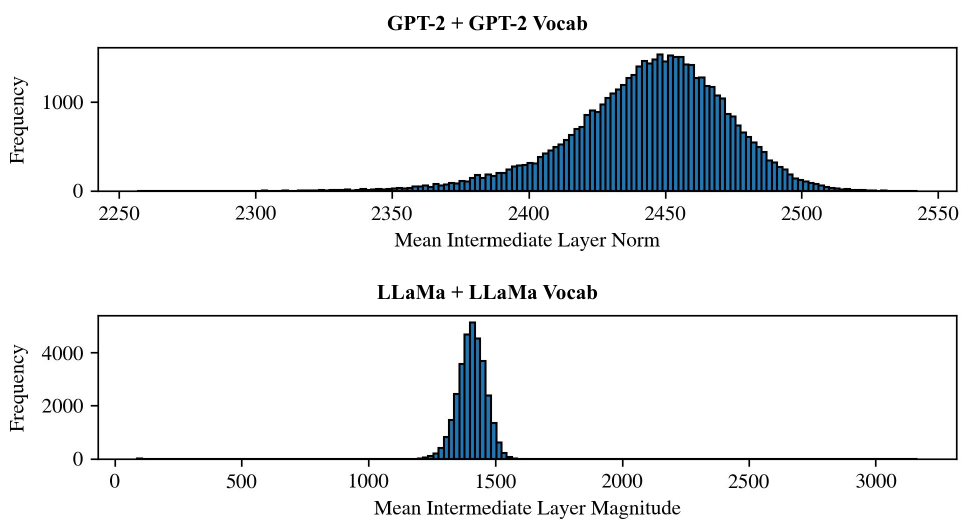}
    \caption{Histograms of the mean intermediate layer latent state norm from each vocab token for both GPT-2 and LLaMa. Each vocab token was input individually into each model as the initial token.}
    \label{fig:singular_norm_histograms}
\end{figure}

When performing dimensionality reduction using methods such as PCA or UMAP on a set of vectors (such as latent states), a few vectors with outlier norms can obscure the structures present between all other vectors in the set by overshadowing them\footnote{In the UMAP case, this is applicable when the chosen metric is sensitive to norm outliers, such as with the Euclidean distance metric. Other metrics, such as cosine distance, would not be affected by norm outliers.}. We analyze the average intermediate layer norms of PG-19 latent states. Figure~\ref{fig:large_norm} shows large spikes in the norm of the 0-th sequence position token for both LLaMa and GPT-2. This phenomenon has been noticed and analyzed in prior works by Sun et al. in 2024~\citep{sun2024massiveactivationslargelanguage} and has functions which include acting as a bias term.

As was similarly described by Sun et al.~\citep{sun2024massiveactivationslargelanguage}, we find that these large spikes in latent norm occur in the LLaMa model at the initial token of a sequence. We find this interesting, as although the LLaMa model uses a \texttt{<BOS>} token at the start of all inputs~\citep{touvron2023llamaopenefficientfoundation, kudo-richardson-2018-sentencepiece}, the spike in latent state norm for the 0-th sequence position is not limited to the appearance of the \texttt{<BOS>} token. Figure~\ref{fig:singular_norm_histograms} shows histograms of the mean norm across intermediate layers for each vocab token in both GPT-2 and LLaMa when given as input to the model as the initial token. As can be seen, the vast majority of LLaMa vocab tokens take on large latent state norms when they are the initial token. We find this property intriguing as, unlike the learned positional embeddings of GPT-2, the \textit{relative} RoPE positional encodings used by LLaMa do not provide the model with as trivial a way to determine whether an embedding belongs to a token in the 0-th sequence position. Nonetheless, tokens in the 0-th sequence position typically have a corresponding latent state in LLaMa with very high norm, indicating that LLaMa has likely developed a more complicated mechanism to detect tokens at the 0-th sequence position.

\begin{figure}[H]
    \centering
    \setlength{\fboxsep}{0pt}
    \includegraphics[width=\textwidth]{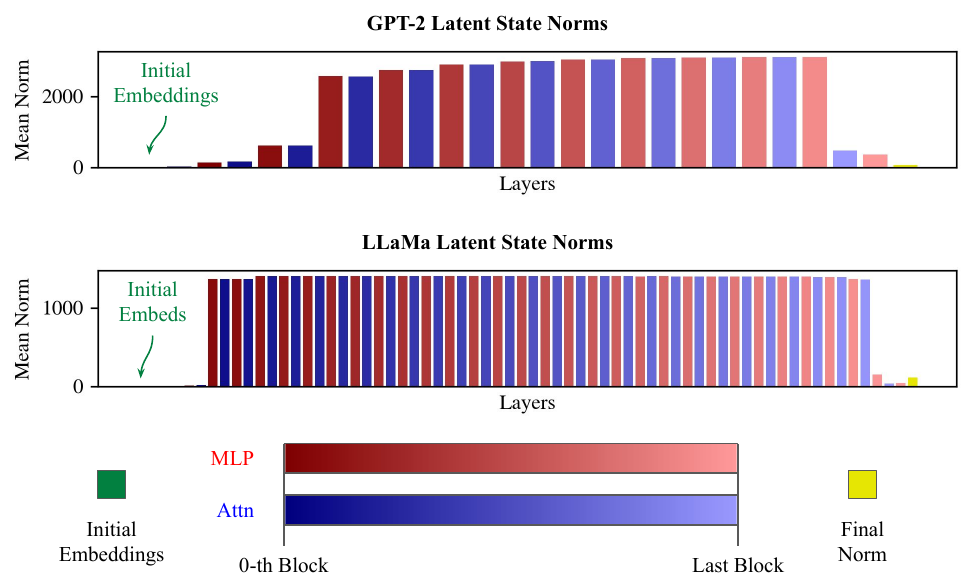}
    \caption{Latent state norms across each layer of both GPT-2 and LLaMa for initial token inputs. For each model, all of its vocab tokens were individually input into the model as the initial token, and the mean norm for each layer was taken over all the vocab tokens.}
    \label{fig:singular_layerwise_norms}
\end{figure}

As was also described by Sun et al.~\citep{sun2024massiveactivationslargelanguage}, the high-norm latent states emerge after a few initial Transformer layers and then reduce in norm in the final layers. Figure~\ref{fig:singular_layerwise_norms} shows the mean norm across all individual vocab tokens (separately input into the model) for each layer of both GPT-2 and LLaMa. As can be seen, the layers in the middle have very high norms, while those at the beginning and end do not.

Due to this observation of high latent state norms in the 0-th (``initial'') sequence position, in subsequent results, we often omit the latent states of the 0-th sequence position when performing dimensionality reduction and visualization so that it does not obscure all details in other latent states.

\subsection{Layerwise Visualizations}
\label{sec:layerwisevisualizations}

\begin{figure}[H]
    \centering
    \setlength{\fboxsep}{0pt}
    \includegraphics[width=\textwidth]{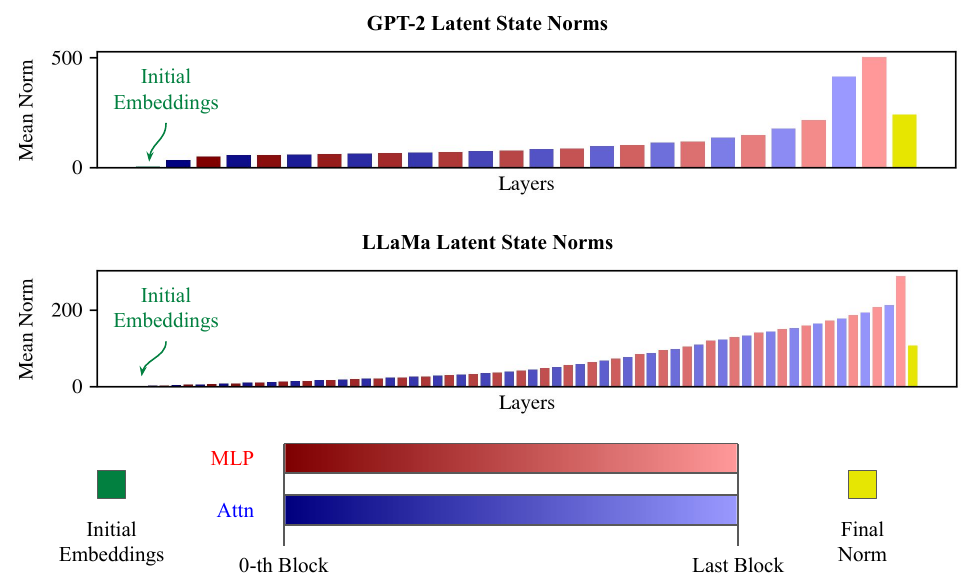}
    \caption{Latent state norms across each layer of both GPT-2 and LLaMa. The norms for each layer were averaged across all samples and sequence positions. Initial token latent states were excluded.}
    \label{fig:layerwise_norms}
\end{figure}

Figure~\ref{fig:layerwise_norms} shows the average norm of each layer of both GPT-2 and LLaMa across sequences of the PG-19 dataset. Initial token latent states are excluded. In contrast with the consistently large norms seen in the initial token latent states of Figure~\ref{fig:singular_layerwise_norms}, the norms of Figure~\ref {fig:layerwise_norms} gradually increase across the layers and are never as large. Furthermore, there are patterns which are consistent across both GPT-2 and LLaMa, such as the large spike in norm in the last block of the Transformer and the drop in norm after the final norm layer.

\begin{figure}[H]
    \centering
    \setlength{\fboxsep}{0pt}
    \includegraphics[width=\textwidth]{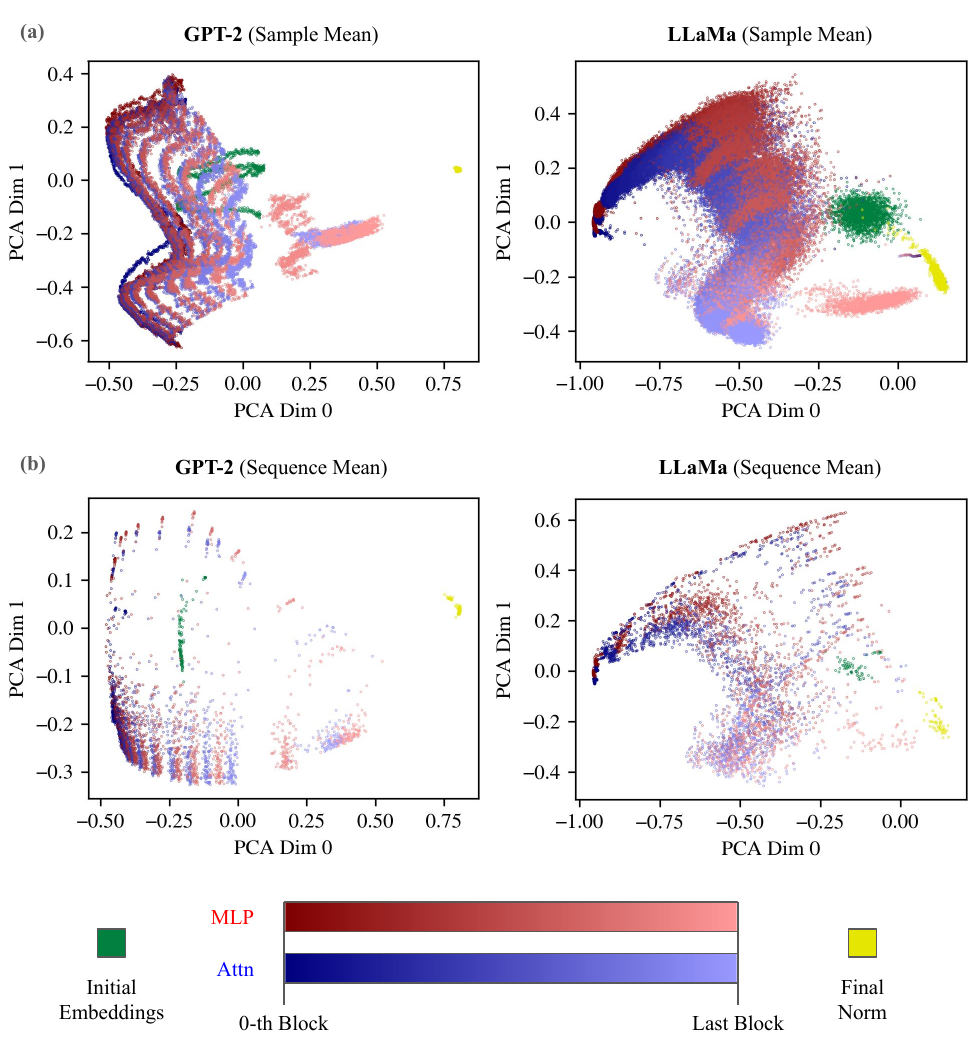}
    \caption{
   PCA visualizations of GPT-2 and LLaMa latent states on the PG-19 dataset. Latent states were converted to unit vectors before any PCA was performed. PCA was fit to all latent states, but was used to transform latent states only after they were averaged across samples/sequence dimensions. The initial token latent states were excluded. (a) Visualization after averaging across the sample dimension. (b) Visualization after averaging across the sequence dimension.}
    \label{fig:layerwise}
\end{figure}

Figure~\ref{fig:layerwise} visualizes, using PCA, the layers of both GPT-2 and LLaMa in latent space. All latent states were converted to unit vectors before any dimensionality reduction or visualization was performed. Otherwise, the higher norms of the later layers tend to overpower all other variability, resulting in an uninteresting visualization. Additionally, while PCA was fit on the full latent state data, it was used to transform the data only after either averaging the latent states over the sequence or sample dimensions in order to reduce visual clutter (otherwise, there can be so many data points that the visualization is not intelligible). Averaging the latent states over the sample dimension removes the variability between samples, leaving the only sources of variability in the latent states being the differences between sequence positions and the differences between layers of the model. In contrast, averaging over the sequence dimension removes any variability/structure associated with the sequence position, thus only leaving the random variability between samples and the differences between layers. As can be seen in figure~\ref{fig:layerwise}, a clear layer-wise progression of latent states is visible in both GPT-2 and LLaMa cases. In the GPT-2 case, the ``wavy'' structure of the positional embeddings is clearly visible in each layer of the sample mean (a) case, while in LLaMa, which uses RoPE, a pattern is less clear. The patterns of sequence position are further analyzed in section~\ref{sec:effects_of_sequence_position}.

\subsection{Attention vs. MLP Signature}
\label{sec:attentionvsmlp}

\begin{figure}[H]
    \centering
    \setlength{\fboxsep}{0pt}
    \includegraphics[width=\textwidth]{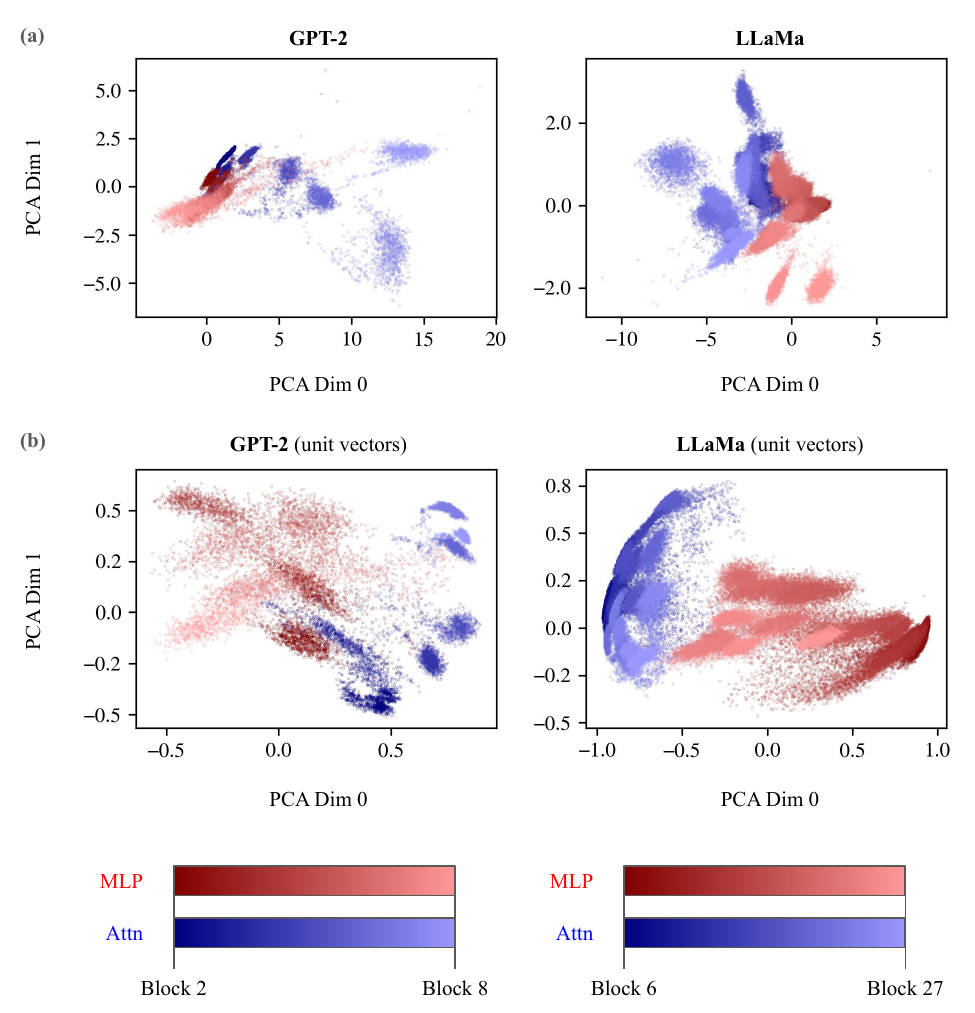}
    \caption{
   PCA visualizations of \textit{pre-add} GPT-2 and LLaMa intermediate layer latent states on the PG-19 dataset. PCA was fit to all latent states but was used to transform latent states after they were averaged across samples. Initial token latent states were excluded. (a) PCA visualizations of GPT-2 and LLaMa intermediate block latent states (b) PCA visualizations of GPT-2 and LLaMa intermediate block latent states after being converted to unit vectors; unit vector data was used for both the fit and transform/visualize stages.}
    \label{fig:attn_vs_mlp_pca}
\end{figure}

\begin{figure}[H]
    \centering
    \setlength{\fboxsep}{0pt}
    \includegraphics[width=\textwidth]{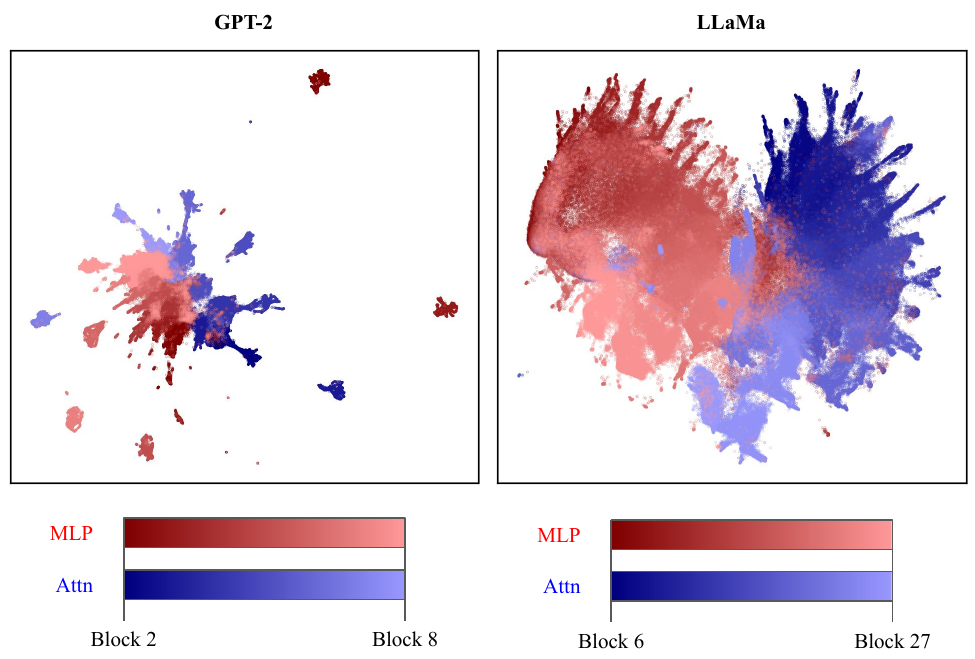}
   \caption{UMAP visualizations of GPT-2 and LLaMa \textit{pre-add} intermediate layer latent states on the PG-19 dataset. For consistency with Figure~\ref{fig:attn_vs_mlp_pca}, initial token latent states are excluded.}
    \label{fig:attn_vs_mlp_umap}
\end{figure}

We notice a striking separation between pre-add latent states from attention components when compared to latent states from MLPs in their dimensionality-reduced visualizations. That is, through many different Transformer blocks, the output of attention components tends to occupy a distinct area in the visualized space when compared to the output of MLPs. To the best of our knowledge, this has not yet been shown in previous studies.

Figures~\ref{fig:attn_vs_mlp_pca} and \ref{fig:attn_vs_mlp_umap} visualize pre-add contributions from both attention components and MLPs in intermediate blocks using samples from PG-19. Figure~\ref{fig:attn_vs_mlp_pca} utilizes PCA dimensionality reduction on the latent states. Although the PCA dimensionality reductions in all subplots of Figure~\ref{fig:attn_vs_mlp_pca} were \textit{fit} on all latent states, the actual \textit{transformation} using the fitted PCA was done on the latent states only after they had all been averaged over the sample dimension to reduce random variability between samples for a clearer visualization. Figure~\ref{fig:attn_vs_mlp_pca}(a) shows the PCA visualizations of both GPT-2 and LLaMa. (b) shows the PCA visualizations of both GPT-2 and LLaMa when all latent states were converted to unit vectors before PCA dimensionality reduction. In both (a) and (b), there is a clear separation of regions occupied by latent states from attention components (blue) versus MLPs (red).

Figure~\ref{fig:attn_vs_mlp_umap} utilized 2D UMAP dimensionality reduction on the latent states. Unlike in the PCA case of Figure~\ref{fig:attn_vs_mlp_pca}, there was no sample averaging of the latent states. Instead, due to the increased computational demands of UMAP (using all latent states is not viable), a random subset of $100\,$k latent states was used to fit UMAP, while a \textit{different} random subset of $500\,$k latent states was transformed and visualized. Similar to as seen in Figure~\ref{fig:attn_vs_mlp_pca}, we can see distinct regions occupied by latent states from attention components (blue) compared to those from MLPs (red). The Linear Representation Hypothesis specifies that features correspond to \textit{directions} in latent space. As a result, we use cosine distance as the metric for UMAP instead of Euclidean distance. We analyze the stability of our UMAP visualizations in appendix~\ref{sec:umap_stability}.

\subsection{Effects of Sequence Position}
\label{sec:effects_of_sequence_position}

We visualize the geometric effects of sequence position on post-add latent states from intermediate blocks of both GPT-2 and LLaMa. PCA was used for dimensionality reduction instead of UMAP, as UMAP does not meaningfully preserve geometric shapes in Euclidean space. We use intermediate layer latent states from PG-19. Latent states were averaged over both samples and layers before dimensionality reduction. This was to eliminate sources of variability between samples or layers, leaving only the structure between sequence positions. Latent states were also converted to unit vectors before dimensionality reduction. This was done for two reasons. First, we found the visualizations looked either nearly identical (GPT-2) or slightly clearer (LLaMa) when latent states were converted to unit vectors. Second, the use of unit vectors allowed us to include the initial token latent state without being concerned that its extreme norm would obscure all other patterns. We provide visualizations that omit the initial token and do not convert latent states to unit vectors as a comparison in appendix~\ref{sec:additional_pos_vis}. For all visualizations, we perform PCA which reduces the latent states to $6$ dimensions and then visualize the $15$ unique pairs of those $6$ dimensions.

\subsubsection{GPT-2 Positional Embeddings}

\begin{figure}[H]
    \centering
    \setlength{\fboxsep}{0pt}
    \includegraphics[width=\textwidth]{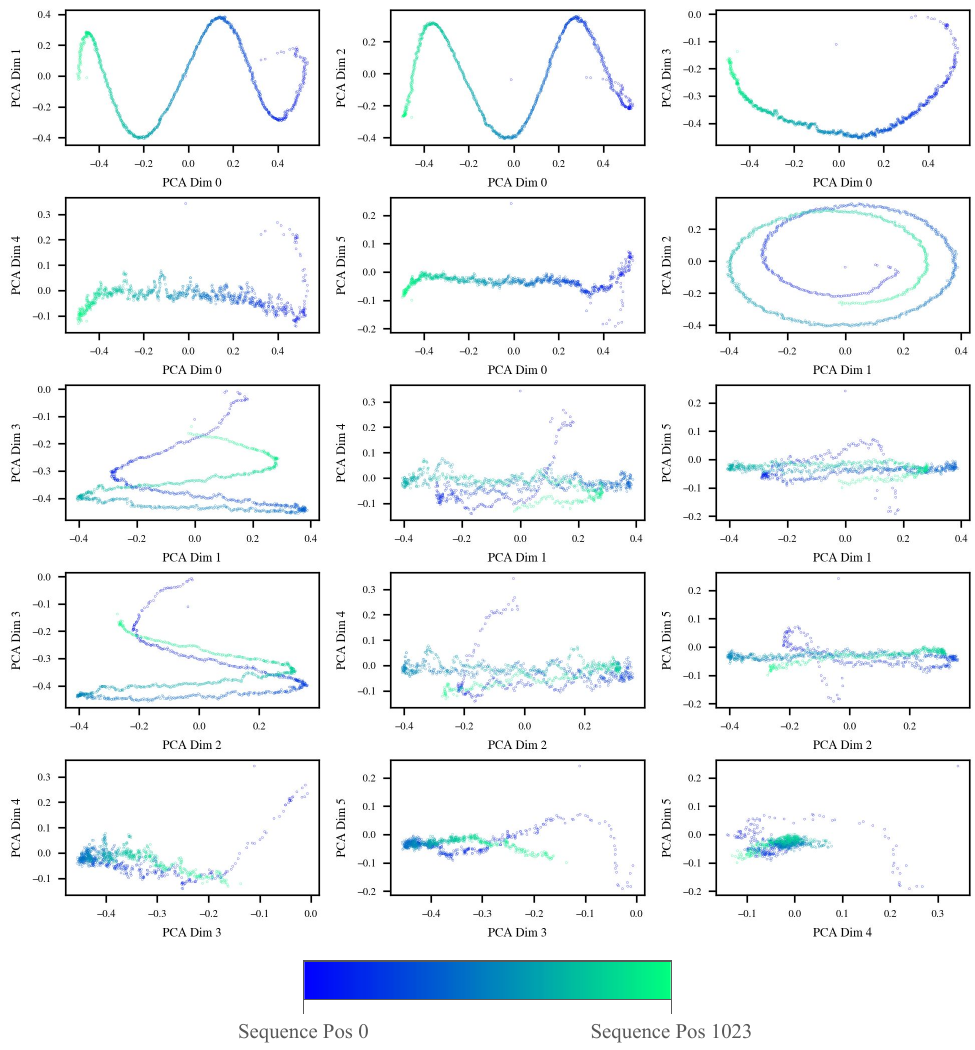}
   \caption{Visualization of post-add latent states from GPT-2 intermediate blocks after conversion to unit vectors, averaged across samples and layers. $15$ plots visualize all unique pairs of $6$ PCA dimensions.}
    \label{fig:gpt2_pos_unit}
\end{figure}

As has already been established in previous works by Scherlis in 2022 and Yedidia in 2023~\citep{scherlis2022exploration, yedidia2023gpt2}, the positional embeddings of GPT-2 form a sort of ``helix'' in high-dimensional space. This can clearly be seen in Figure~\ref{fig:gpt2_pos_unit}. We also notice that a clear pattern across the sequence positions remains prominent for all combinations of PCA dimensions, demonstrating the high dimensionality of the geometric pattern formed by the GPT-2 positional embeddings.

\subsubsection{LLaMa Positional Encodings}

\begin{figure}[H]
    \centering
    \setlength{\fboxsep}{0pt}
    \includegraphics[width=\textwidth]{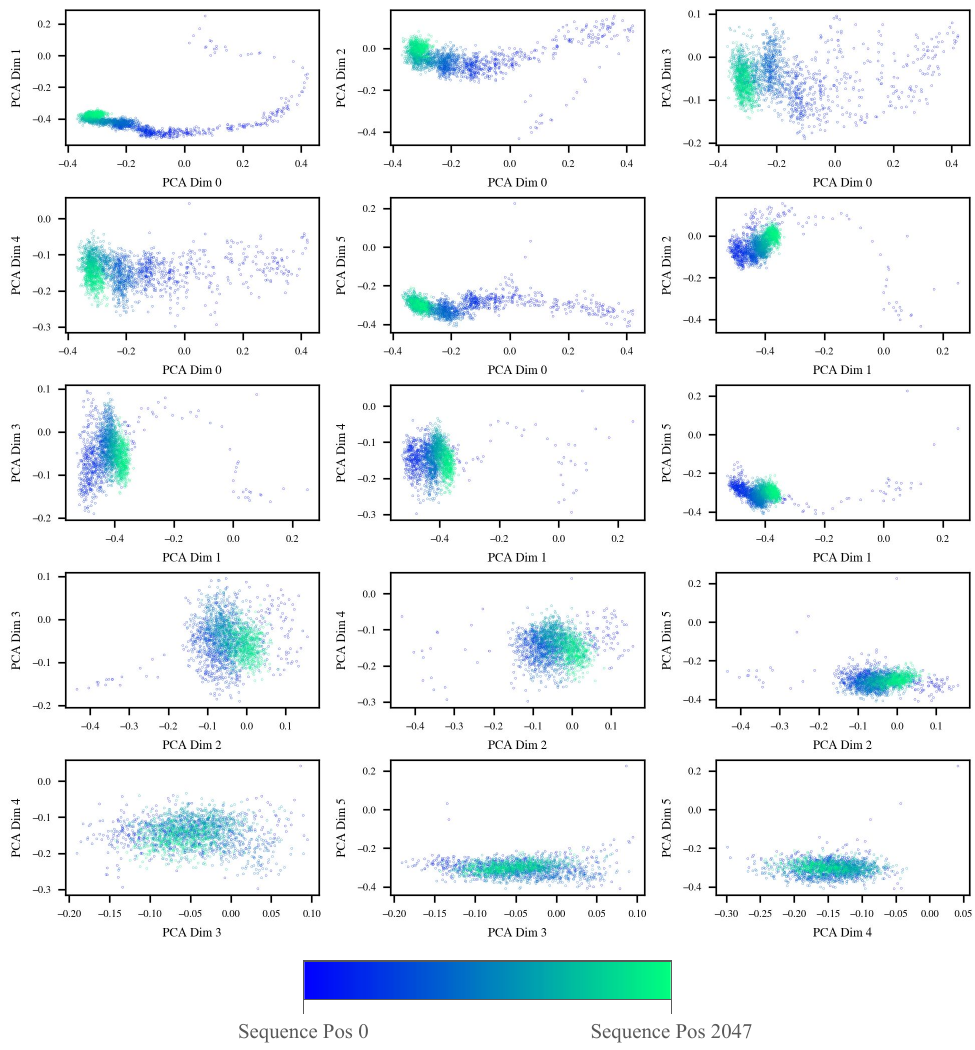}
   \caption{Visualization of post-add latent states from LLaMa intermediate blocks after conversion to unit vectors, averaged across samples and layers. $15$ plots visualize all unique pairs of $6$ PCA dimensions.}
    \label{fig:llama_pos_unit}
\end{figure}

In Figure~\ref{fig:llama_pos_unit}, we see a clear geometric pattern formed between sequence positions in subplots visualizing pairs of earlier PCA dimensions. In the first $3$ rows of subplots, there appears to be a long "tail" of initial sequence positions which then leads into a dense cloud of latent states at later sequence positions. In subplots visualizing pairs of later PCA dimensions (the final $2$ rows), this pattern disappears, and the latent states instead form a relatively uniform ``cloud'' with no significant pattern formed by the sequence positions. This may be an indication that the geometric pattern formed by the sequence positions is relatively low-dimensional, as it quickly disappears within the first few PCA dimensions.

Due to the inherent nature of RoPE applying only \textit{within} self-attention heads, it is not straightforward to fully separate the effects of the RoPE-augmented attention heads on latent states from other factors such as the content of the input tokens themselves. This is unlike how it is possible to separate the learned positional embeddings of GPT-2 from any other context by simply analyzing the learned positional embeddings themselves. As such, it is not possible to determine via these visualizations alone whether the geometric patterns observed in Figure~\ref{fig:llama_pos_unit} are a result of RoPE, the relative convergence of features within contributions from self-attention heads as the number of previous tokens increases, or both. We leave interpreting the sequence-wise latent state geometric patterns of RoPE models to future research.

\section{Conclusion}

We present a study on visualizing the internal representations of Transformer-based language models. Through systematic analysis of GPT-2 and LLaMa, we demonstrate how dimensionality reduction techniques can reveal geometric patterns that show how these models organize and process information. Our experiments highlight several notable phenomena. Most significantly, we identify a persistent geometric separation between attention and MLP component outputs (section~\ref{sec:attentionvsmlp}), a pattern that appears consistent across the tested model architectures and has not been previously documented to our knowledge. We also noted the high norm of latent states at the initial sequence position (section~\ref{sec:first_pos_large_norm}), a phenomenon that extends beyond special tokens like \texttt{<BOS>} to many vocabulary tokens in LLaMa despite its use of relative position encodings. Additionally, we visualized the layerwise evolution of latent states (section~\ref{sec:layerwisevisualizations}) and the geometric effects of sequence position (section~\ref{sec:effects_of_sequence_position}). Ultimately, we add to the growing body of interpretability research by motivating further work into analyzing feature geometry. Future work can help further our understanding of the dynamics of latent states within Transformer models across layers, models, and experimental conditions. By deepening our understanding of how these geometric structures emerge and behave, we move closer to principled, reliable interpretability methods capable of guiding the development of more transparent and understandable models.

\section*{Acknowledgments}
We thank the University of Virginia Research Computing and the Department of Computer Science for providing the computational resources that made this research possible.

\bibliographystyle{unsrt}  
\bibliography{references} 

@article{elhage2021mathematical,
   title={A Mathematical Framework for Transformer Circuits},
   author={Elhage, Nelson and Nanda, Neel and Olsson, Catherine and Henighan, Tom and Joseph, Nicholas and Mann, Ben and Askell, Amanda and Bai, Yuntao and Chen, Anna and Conerly, Tom and DasSarma, Nova and Drain, Dawn and Ganguli, Deep and Hatfield-Dodds, Zac and Hernandez, Danny and Jones, Andy and Kernion, Jackson and Lovitt, Liane and Ndousse, Kamal and Amodei, Dario and Brown, Tom and Clark, Jack and Kaplan, Jared and McCandlish, Sam and Olah, Chris},
   year={2021},
   journal={Transformer Circuits Thread},
   note={https://transformer-circuits.pub/2021/framework/index.html}
}

@article{elhage2022superposition,
   title={Toy Models of Superposition},
   author={Elhage, Nelson and Hume, Tristan and Olsson, Catherine and Schiefer, Nicholas and Henighan, Tom and Kravec, Shauna and Hatfield-Dodds, Zac and Lasenby, Robert and Drain, Dawn and Chen, Carol and Grosse, Roger and McCandlish, Sam and Kaplan, Jared and Amodei, Dario and Wattenberg, Martin and Olah, Christopher},
   year={2022},
   journal={Transformer Circuits Thread},
   note={https://transformer-circuits.pub/2022/toy_model/index.html}
}

@misc{nanda2023factfinding,
  title={Fact Finding: Attempting to Reverse-Engineer Factual Recall on the Neuron Level},
  url={https://www.alignmentforum.org/posts/iGuwZTHWb6DFY3sKB/fact-finding-attempting-to-reverse-engineer-factual-recall},
  journal={Alignment Forum},
  author={Nanda, Neel and Rajamanoharan, Senthooran and Kramar, Janos and Shah, Rohin},
  year={2023},
  month={Dec}
}

@misc{park2024linearrepresentationhypothesisgeometry,
      title={The Linear Representation Hypothesis and the Geometry of Large Language Models}, 
      author={Kiho Park and Yo Joong Choe and Victor Veitch},
      year={2024},
      eprint={2311.03658},
      archivePrefix={arXiv},
      primaryClass={cs.CL},
      url={https://arxiv.org/abs/2311.03658}, 
}

@inproceedings{mikolov-etal-2013-linguistic,
    title = "Linguistic Regularities in Continuous Space Word Representations",
    author = "Mikolov, Tomas  and
      Yih, Wen-tau  and
      Zweig, Geoffrey",
    editor = "Vanderwende, Lucy  and
      Daum{\'e} III, Hal  and
      Kirchhoff, Katrin",
    booktitle = "Proceedings of the 2013 Conference of the North {A}merican Chapter of the Association for Computational Linguistics: Human Language Technologies",
    month = {jun},
    year = "2013",
    address = "Atlanta, Georgia",
    publisher = "Association for Computational Linguistics",
    url = "https://aclanthology.org/N13-1090/",
    pages = "746--751"
}

@misc{engels2025languagemodelfeaturesonedimensionally,
      title={Not All Language Model Features Are One-Dimensionally Linear}, 
      author={Joshua Engels and Eric J. Michaud and Isaac Liao and Wes Gurnee and Max Tegmark},
      year={2025},
      eprint={2405.14860},
      archivePrefix={arXiv},
      primaryClass={cs.LG},
      url={https://arxiv.org/abs/2405.14860}, 
}

@misc{vaswani2023attentionneed,
      title={Attention Is All You Need}, 
      author={Ashish Vaswani and Noam Shazeer and Niki Parmar and Jakob Uszkoreit and Llion Jones and Aidan N. Gomez and Lukasz Kaiser and Illia Polosukhin},
      year={2023},
      eprint={1706.03762},
      archivePrefix={arXiv},
      primaryClass={cs.CL},
      url={https://arxiv.org/abs/1706.03762}, 
}

@misc{scherlis2022exploration,
      title={An Exploration of GPT-2's Embedding Weights}, 
      author={Adam Scherlis},
      year={2022},
      month={Dec},
      howpublished={\url{https://www.lesswrong.com/posts/BMghmAxYxeSdAteDc/an-exploration-of-gpt-2-s-embedding-weights}},
}

@misc{yedidia2023gpt2,
      title={GPT-2's Positional Embedding Matrix is a Helix}, 
      author={Adam Yedidia},
      year={2023},
      month={Jul},
      howpublished={\url{https://www.lesswrong.com/posts/qvWP3aBDBaqXvPNhS/gpt-2-s-positional-embedding-matrix-is-a-helix}},
}

@misc{mcinnes2020umapuniformmanifoldapproximation,
      title={UMAP: Uniform Manifold Approximation and Projection for Dimension Reduction}, 
      author={Leland McInnes and John Healy and James Melville},
      year={2020},
      eprint={1802.03426},
      archivePrefix={arXiv},
      primaryClass={stat.ML},
      url={https://arxiv.org/abs/1802.03426}, 
}

@misc{shlens2014tutorialprincipalcomponentanalysis,
      title={A Tutorial on Principal Component Analysis}, 
      author={Jonathon Shlens},
      year={2014},
      eprint={1404.1100},
      archivePrefix={arXiv},
      primaryClass={cs.LG},
      url={https://arxiv.org/abs/1404.1100}, 
}

@misc{raschka2020machinelearningpythonmain,
      title={Machine Learning in Python: Main developments and technology trends in data science, machine learning, and artificial intelligence}, 
      author={Sebastian Raschka and Joshua Patterson and Corey Nolet},
      year={2020},
      eprint={2002.04803},
      archivePrefix={arXiv},
      primaryClass={cs.LG},
      url={https://arxiv.org/abs/2002.04803}, 
}

@misc{sun2024massiveactivationslargelanguage,
      title={Massive Activations in Large Language Models}, 
      author={Mingjie Sun and Xinlei Chen and J. Zico Kolter and Zhuang Liu},
      year={2024},
      eprint={2402.17762},
      archivePrefix={arXiv},
      primaryClass={cs.CL},
      url={https://arxiv.org/abs/2402.17762}, 
}

@misc{touvron2023llamaopenefficientfoundation,
      title={LLaMA: Open and Efficient Foundation Language Models}, 
      author={Hugo Touvron and Thibaut Lavril and Gautier Izacard and Xavier Martinet and Marie-Anne Lachaux and Timothée Lacroix and Baptiste Rozière and Naman Goyal and Eric Hambro and Faisal Azhar and Aurelien Rodriguez and Armand Joulin and Edouard Grave and Guillaume Lample},
      year={2023},
      eprint={2302.13971},
      archivePrefix={arXiv},
      primaryClass={cs.CL},
      url={https://arxiv.org/abs/2302.13971}, 
}

@inproceedings{kudo-richardson-2018-sentencepiece,
    title = "{S}entence{P}iece: A simple and language independent subword tokenizer and detokenizer for Neural Text Processing",
    author = "Kudo, Taku  and
      Richardson, John",
    editor = "Blanco, Eduardo  and
      Lu, Wei",
    booktitle = "Proceedings of the 2018 Conference on Empirical Methods in Natural Language Processing: System Demonstrations",
    month = {nov},
    year = "2018",
    address = "Brussels, Belgium",
    publisher = "Association for Computational Linguistics",
    url = "https://aclanthology.org/D18-2012/",
    doi = "10.18653/v1/D18-2012",
    pages = "66--71"
}

@article{radford2019language,
  title={Language Models are Unsupervised Multitask Learners},
  author={Radford, Alec and Wu, Jeff and Child, Rewon and Luan, David and Amodei, Dario and Sutskever, Ilya},
  year={2019}
}

@misc{rae2019compressivetransformerslongrangesequence,
      title={Compressive Transformers for Long-Range Sequence Modelling}, 
      author={Jack W. Rae and Anna Potapenko and Siddhant M. Jayakumar and Timothy P. Lillicrap},
      year={2019},
      eprint={1911.05507},
      archivePrefix={arXiv},
      primaryClass={cs.LG},
      url={https://arxiv.org/abs/1911.05507}, 
}

@misc{sun2025transformerlayerspainters,
      title={Transformer Layers as Painters}, 
      author={Qi Sun and Marc Pickett and Aakash Kumar Nain and Llion Jones},
      year={2025},
      eprint={2407.09298},
      archivePrefix={arXiv},
      primaryClass={cs.CL},
      url={https://arxiv.org/abs/2407.09298}, 
}

@misc{lad2025remarkablerobustnessllmsstages,
      title={The Remarkable Robustness of LLMs: Stages of Inference?}, 
      author={Vedang Lad and Jin Hwa Lee and Wes Gurnee and Max Tegmark},
      year={2025},
      eprint={2406.19384},
      archivePrefix={arXiv},
      primaryClass={cs.LG},
      url={https://arxiv.org/abs/2406.19384}, 
}

@article{JMLR:v9:vandermaaten08a,
  author  = {Laurens van der Maaten and Geoffrey Hinton},
  title   = {Visualizing Data using t-SNE},
  journal = {Journal of Machine Learning Research},
  year    = {2008},
  volume  = {9},
  number  = {86},
  pages   = {2579--2605},
  url     = {http://jmlr.org/papers/v9/vandermaaten08a.html}
}

@misc{su2023roformerenhancedtransformerrotary,
      title={RoFormer: Enhanced Transformer with Rotary Position Embedding}, 
      author={Jianlin Su and Yu Lu and Shengfeng Pan and Ahmed Murtadha and Bo Wen and Yunfeng Liu},
      year={2023},
      eprint={2104.09864},
      archivePrefix={arXiv},
      primaryClass={cs.CL},
      url={https://arxiv.org/abs/2104.09864}, 
}

\section*{Glossary}
\label{sec:glossary}
\noindent
\textbf{0-Based Indexing} — A convention where counting begins at zero, used for numbering Transformer blocks and sequence positions (e.g., “block 0” or “sequence position 0”).  
\textit{Section~\ref{subsec:Transformers}}

\vspace{0.5em}
\noindent
\textbf{Attention Component} — A Transformer component that computes relationships and transmits information between tokens via self-attention, enabling context-dependent representations.  
\textit{Section~\ref{subsec:Transformers}}

\vspace{0.5em}
\noindent
\textbf{Feature Geometry} — The structure and organization of learned representations within high-dimensional latent space, often analyzed through dimensionality reduction.  
\textit{Section~\ref{sec:introduction}}

\vspace{0.5em}
\noindent
\textbf{Initial Token} — The first token in a sequence (position 0).  
\textit{Section~\ref{sec:first_pos_large_norm}}

\vspace{0.5em}
\noindent
\textbf{Intermediate Layers/Blocks} — The middle portion of a Transformer, as defined in the Experimental Setup section.  
\textit{Section~\ref{subsec:ExperimentalSetup}}

\vspace{0.5em}
\noindent
\textbf{Layer Dimension} — The dimension of our generated dataset that indexes different layers/components across the Transformer's depth.  
\textit{Section~\ref{subsec:pipeline}}

\vspace{0.5em}
\noindent
\textbf{Layers / Components} — The key submodules within each Transformer block: \textit{normalization layers}, \textit{attention}, and \textit{MLP}.  
\textit{Section~\ref{subsec:Transformers}}

\vspace{0.5em}
\noindent
\textbf{Latent Space} — The high-dimensional vector space in which internal model representations (latent states) reside.  
\textit{Section~\ref{subsec:Transformers}}

\vspace{0.5em}
\noindent
\textbf{Learned Positional Embeddings} — A method of generating positional embeddings by learning high-dimensional additive embedding vectors.  
\textit{Section~\ref{subsec:positionalencodings}}

\vspace{0.5em}
\noindent
\textbf{Linear Representation Hypothesis (LRH)} — The hypothesis that high-level features in language models correspond to approximately linear directions in representation space.  
\textit{Section~\ref{subsec:lrh}}

\vspace{0.5em}
\noindent
\textbf{MLP (Multilayer Perceptron)} — The feed-forward component of a Transformer block.  
\textit{Section~\ref{subsec:Transformers}}

\vspace{0.5em}
\noindent
\textbf{Norm} — The L2 norm (Euclidean length/magnitude) of a vector.  
\textit{Section~\ref{sec:first_pos_large_norm}}

\vspace{0.5em}
\noindent
\textbf{Normalization Layer} — A component (e.g., LayerNorm, RMSNorm) that stabilizes activations by rescaling and centering inputs; may appear before or after components in pre-norm or post-norm designs.  
\textit{Section~\ref{subsec:Transformers}}

\vspace{0.5em}
\noindent
\textbf{PCA (Principal Component Analysis)} — A linear dimensionality reduction technique.  
\textit{Section~\ref{subsec:dimreduct}}

\vspace{0.5em}
\noindent
\textbf{Post-Add Latent States} — Latent states captured after an attention or MLP output has been added back into the residual stream, reflecting updated representations.  
\textit{Section~\ref{subsec:Transformers}}

\vspace{0.5em}
\noindent
\textbf{Pre-Add Latent States} — Output of an attention or MLP component before it is added back into the residual stream.  
\textit{Section~\ref{subsec:Transformers}}

\vspace{0.5em}
\noindent
\textbf{Residual Stream} — The central communication channel formed by skip connections in the Transformer.  
\textit{Section~\ref{subsec:Transformers}}

\vspace{0.5em}
\noindent
\textbf{RoPE (Rotary Positional Encoding)} — A method for injecting position information through rotating attention query and key vectors in the attention mechanism.  
\textit{Section~\ref{subsec:positionalencodings}}

\vspace{0.5em}
\noindent
\textbf{Samples} — Individual text inputs or tokenized passages passed through the model as batches to produce latent states for analysis.  
\textit{Section~\ref{subsec:pipeline}}

\vspace{0.5em}
\noindent
\textbf{Sample Dimension} — The dimension of our generated latent state dataset that indexes different input samples.  
\textit{Section~\ref{subsec:pipeline}}

\vspace{0.5em}
\noindent
\textbf{Sequence Dimension} — The dimension of our generated latent state dataset that indexes different token positions within a sequence.  
\textit{Section~\ref{subsec:pipeline}}

\vspace{0.5em}
\noindent
\textbf{Skip Connections} — Pathways that add each block’s input back to its output, enabling a shared representation space across layers.  
\textit{Section~\ref{subsec:Transformers}}

\vspace{0.5em}
\noindent
\textbf{Unembedding Layer} — The final linear layer that projects the model’s last hidden state back into the vocabulary space.  
\textit{Section~\ref{subsec:Transformers}}

\vspace{0.5em}
\noindent
\textbf{UMAP (Uniform Manifold Approximation and Projection)} — A nonlinear dimensionality reduction technique.  
\textit{Section~\ref{subsec:dimreduct}}

\section{Appendix}

\subsection{Additional Sequence Position Visualizations}
\label{sec:additional_pos_vis}

\begin{figure}[H]
    \centering
    \setlength{\fboxsep}{0pt}
    \includegraphics[width=\textwidth]{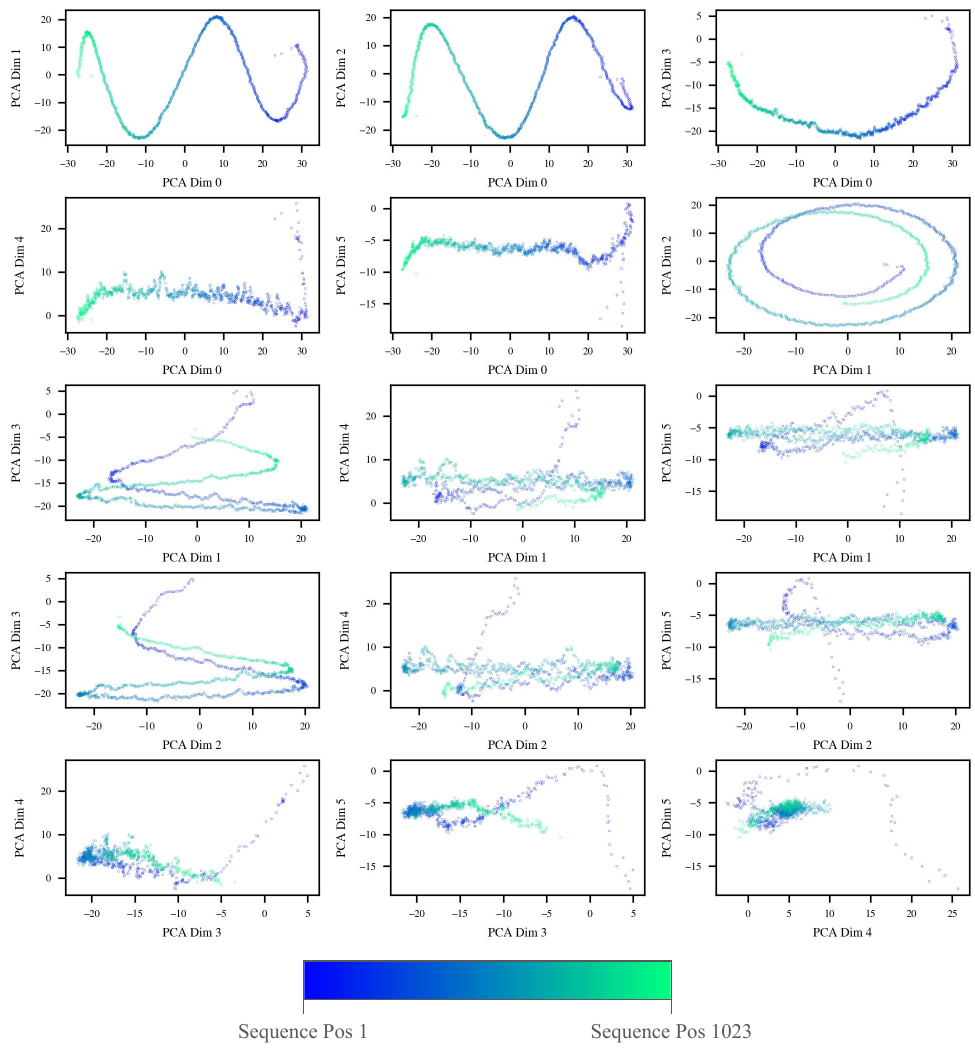}
   \caption{Visualization of post-add latent states from GPT-2 intermediate blocks, averaged across samples and layers. $15$ plots visualize all unique pairs of $6$ PCA dimensions.}
    \label{fig:gpt2_pos}
\end{figure}

\begin{figure}[H]
    \centering
    \setlength{\fboxsep}{0pt}
    \includegraphics[width=\textwidth]{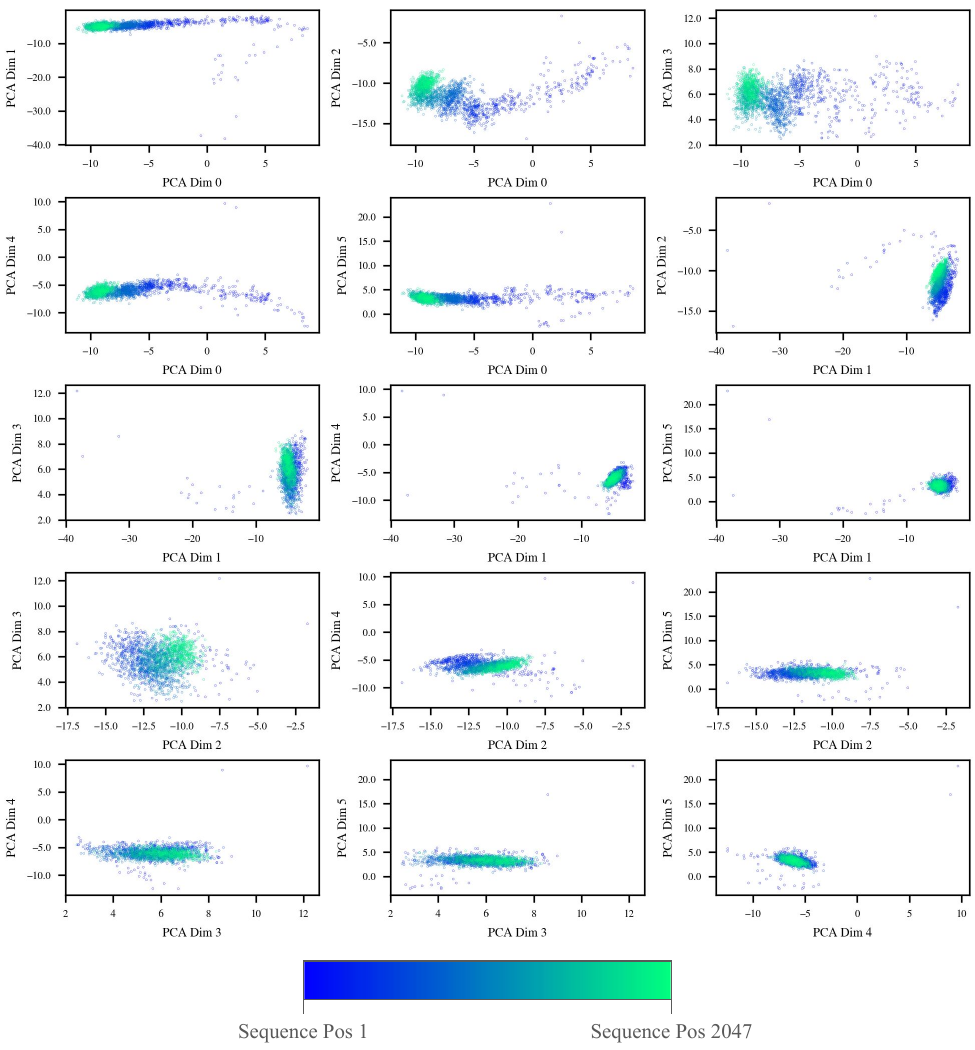}
   \caption{Visualization of post-add latent states from LLaMa intermediate blocks, averaged across samples and layers. $15$ plots visualize all unique pairs of $6$ PCA dimensions.}
    \label{fig:llama_pos}
\end{figure}

\subsection{Stability of UMAP Visualizations}
\label{sec:umap_stability}

\textit{This section discusses the stability of the UMAP visualizations presented in section~\ref{sec:attentionvsmlp}. It is a winding, exploratory rabbit-hole. We included it to discuss the concerns brought up by a reviewer during peer-review regarding the stability of UMAP.}

UMAP is stochastic and, at least in our case, required some largely heuristics-driven and trial-and-error based hyperparameter tuning.

We chose to use two separately sampled random sets of data points for fit and then transformation to minimize the risk of overfitting affecting our results. The particular numbers of data points chosen - 100k for fit and 500k for transform - were simply a result of practical compute and time considerations.

Since UMAP is non-deterministic, different runs will produce different outcomes. As such, in order to ensure that our results were not simply spurious, we can run the process multiple times to see whether the result will still contain the same patterns.

We find that the UMAP visualizations of GPT-2 hidden states are very stable, consistently producing the same, elegant patterns. This is also largely true of the visualizations of LLaMa hidden states. The clear separation of attention and MLP hidden states are always present. However, we also find that the visualized LLaMa hidden states are sometimes susceptible to random and sparse outlier points that visually compact the vast majority of hidden states into a smaller area on the visualization.

\begin{figure}[H]
    \centering
    \setlength{\fboxsep}{0pt}
    \includegraphics[width=\textwidth]{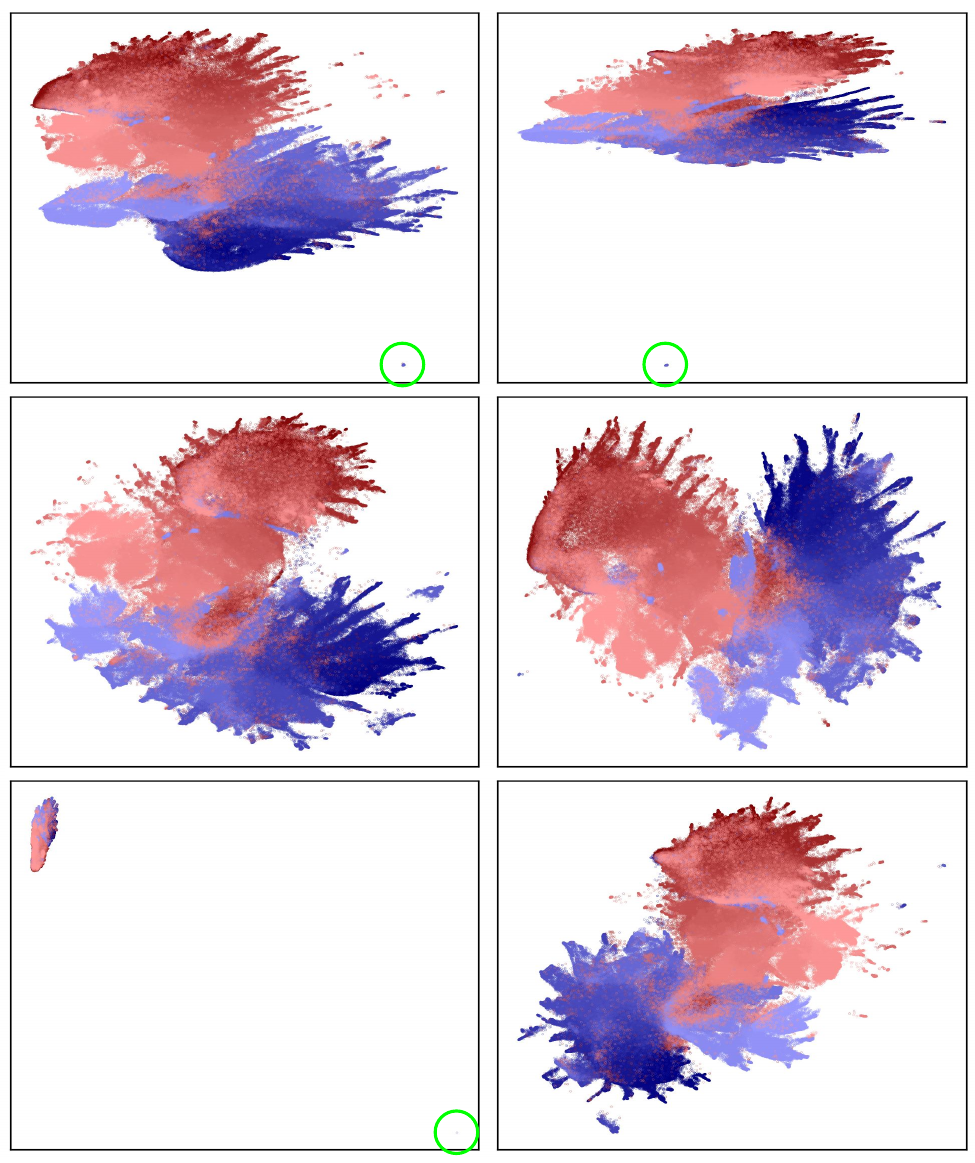}
   \caption{$6$ random UMAP visualizations of LLaMa hidden states. Outliers are circled in green. The top 2 contain outliers that moderately compact the vast majority of data points. The bottom-left visualization has its main cluster of data points severely crushed by distant outliers. The remaining 3 visualizations are not affected by outliers.}
    \label{fig:llama_umap_outliers}
\end{figure}

Since we used cosine similarity as the UMAP metric, we know that large activations are \textit{not} the culprit. Instead, these outliers must simply point in a direction that is very different from the majority of points. Either that, or they are some sort of artifact of a suboptimally-configured UMAP.

One of the simpler reasons that directional outliers may occur is in the case where there are hidden states that have a near-zero magnitude, since these can essentially point in a random, meaningless direction. However, we find that this is not the case. Out of all LLaMa hidden states we considered for visualization (after initial post-processing dimensionality reduction via PCA from $4096$ to $512$ dims), the minimum Euclidean length is $0.87$ while the median is $5.73$; this minimum length is far too great and eliminates near-zero length as a possible culprit of the directional outliers.

We do not find that transforming the \textit{same} data points that were used to fit the UMAP (in comparison with using separately sampled sets) has a noticeable effect on the occurrence of outliers.

We proceed to attempt to provide some interpretation on the source of these outliers. Out of the $64$ text samples that comprised the hidden states we used, we manually examined the first $15$. We found outliers present in 2 out of these $15$ samples (samples 12 and 14). Out of all of the $90{,}068$ hidden states that comprised sample 12, we found that only 1 was an outlier: the hidden state output by the MLP of block 20 at sequence position 1596, which corresponds to the final token \texttt{'.'} in the following sequence of tokens that discusses what seems to be a delectable experience:

\begin{verbatim}
...

'▁Mix', '▁well', '▁with', '▁a', '▁small', '▁glass',
'▁of', '<0x0A>', 'br', 'andy', '.', '▁Let', '▁free',
'ze', '▁and', '▁serve', '▁in', '▁small', '▁glass',
'es', '.', '<0x0A>', '<0x0A>', '<0x0A>', '1', '2',
'.--', 'Russ', 'ian', '▁P', 'anc', 'akes', '.'

...
\end{verbatim}

In sample 14, we find 137 outliers scattered relatively evenly throughout the passage. All are from the output of the attention of block 19. 134 correspond to the token \texttt{'▁|'}, while 3 correspond to the token \texttt{'-{}-+'}. We note that there exist instances of these same tokens that are \textit{not} outliers.

We also measure the cosine similarity between the singular outlier in sample 12 and one of the \texttt{'▁|'} outliers from sample 14. We find a very low cosine similarity of $\sim$0.08. In contrast, cosine similarity between \texttt{'▁|'} outliers in sample 14 is around 0.9. Thus, there is low similarity between the outliers of these two samples, despite the fact that, when reduced with UMAP, the points are essentially overlapping.

We do not attempt to manually identify the remaining outliers in other samples, although we note they do exist. From the two samples we analyze that do contain outliers, we see that they correspond to tokens related to punctuation or formatting and that, given a single sample, they occur only in one layer throughout the entire model.

All of the outliers analyzed above were from a single, \textit{particular} UMAP fit. We do not know if another instance of the stochastic UMAP will produce different outliers.

We do find that fitting on more data points increases the probability and intensity of outliers. For example, fitting on 100k data points produces many visualizations with no outliers, but fitting on 500k data points nearly always produces extreme outliers that compact the majority of data points into a small cluster. Notably, in both cases we transform and visualize 500k data points after fitting. We do not have a great explanation for this. Perhaps there are just a few data points in the 5.8 million total that, when sampled into the set used to fit UMAP, will result in the phenomenon of outliers being present. Thus, it is more likely that none of these data points are sampled when only 100k data points are used rather than 500k. However, if there are hundreds of these ``outlier causing'' data points, then it is very unlikely that even sampling 100k data points out of 5.8 million would miss all of them.

As an alternative to discovering the source of our outliers, we tried to tune the UMAP fitting process to eliminate the phenomenon entirely. We spent a full day playing with various hyperparameters, however, we were not able to eliminate the issue entirely. Despite this, we reiterate that, even with the occasional presence of significant outliers, the primary pattern of interest - the distinction between attention and MLP hidden states - are still \textit{always} evident in the main cluster of visualized data points. As a result, our primary conclusion is still stable across UMAP runs.

\end{document}